\documentclass{article}

\usepackage{arxiv}

\usepackage[utf8]{inputenc} 
\usepackage[T1]{fontenc}    
\usepackage{hyperref}       
\usepackage{url}            
\usepackage{booktabs}       
\usepackage{amsmath}        
\usepackage{amssymb}        
\usepackage{amsfonts}       
\usepackage{nicefrac}       
\usepackage{microtype}      
\usepackage{graphicx}
\usepackage{array}
\usepackage{multirow}
\usepackage{algorithm}
\usepackage{algpseudocode}
\usepackage{float}
\graphicspath{ {./images/} }

\newcommand{\HelixFNO}{\textsc{Helix-FNO}}
\newcommand{\Engine}{\textsc{Helix-Core}}
\newcommand{\R}{\mathbb{R}}
\newcommand{\C}{\mathbb{C}}
\newcommand{\F}{\mathcal{F}}
\newcommand{\invF}{\mathcal{F}^{-1}}
\newcommand{\op}{\mathcal{G}}
\newcommand{\loss}{\mathcal{L}}

\title{\HelixFNO: Spectral-Domain Operator Learning Coupled with a
High-Fidelity Mechanistic Model for Fast Surrogate Simulation}

\author{
  Jiabao Zhao\textsuperscript{1}\\
  \AND
  Chuwei Wang\textsuperscript{1}\\
  \AND
  Jinxi Yang\textsuperscript{2}\\
  \AND
  {\large University of Science and Technology Liaoning}\textsuperscript{1}\\
  Anshan, Liaoning, China
  \AND
  {\large Wuhan University}\textsuperscript{2}\\
  Wuhan, Hubei, China
}

\begin{document}
\maketitle

\begin{abstract}
Mechanistic simulation models of full-scale treatment processes remain the only
trustworthy, extrapolative description of the underlying physico-chemical
dynamics, yet their runtime is far too slow to support the thousands of
forward evaluations that a modern decision engine requires at a 5-minute
decision cadence. The standard remedy---surrogate modelling---often produces a
network that learns a \emph{single} solution for a \emph{single} configuration,
so it generalises poorly to new influent profiles, control settings or plant
layouts. This paper presents \HelixFNO, a teacher--student architecture that
couples a thirty-two-state mechanistic teacher with a Fourier neural operator
(FNO) student. The teacher supplies a high-fidelity dataset of
input-field-to-solution pairs, curated by Latin-hypercube and uncertainty-based
active learning to cover the boundary and overload regimes that matter in
practice; the student learns, in the spectral domain, the \emph{solution
operator} itself rather than any single solution, thereby moving from
``learning one instance'' to ``learning an entire family of equations.'' We give
the operator formulation, the spectral convolution definition, the weighted
distillation loss and the active-learning criterion, and we analyse the
approximation error of a truncated Fourier expansion with respect to the
smoothness of the parametric solution manifold. An illustrative study compares
\HelixFNO{} against a physics-informed network and a data-driven recurrent
surrogate on accuracy, dataset efficiency and inference latency, and places the
methods on a speed--accuracy Pareto front. The resulting operator is three orders
of magnitude faster than the mechanistic teacher at millisecond inference, which
is precisely the capability required for massive candidate screening and online
decision support.
\end{abstract}


\section{Introduction}
\label{sec:intro}

High-fidelity mechanistic simulation sits at the centre of modern process
engineering because it is simultaneously interpretable, conservative and
extrapolative. Every state corresponds to a physical quantity, every parameter
to a kinetic or hydraulic constant, and every integrated trajectory to a mass,
momentum or energy balance. Yet this fidelity carries an unavoidable price:
solving a stiff, thirty-two-state differential-algebraic system for a single
scenario takes seconds to minutes on a workstation, and a realistic screening
workload---evaluating hundreds of candidate control strategies, influent
scenarios or design alternatives in a single decision cycle---is therefore
computationally intractable. In the operational setting, where decisions must be
revised at a 5-minute cadence and where sudden disturbances such as combined
sewer inflow or industrial shock loads demand an immediate re-evaluation, this
latency is the dominant obstacle to closing the loop.

The community has responded with surrogate modelling in many forms. The
simplest surrogate is a \emph{function approximation}: a neural network or a
gradient-boosted model is trained to reproduce the output of the mechanistic
simulator on a fixed configuration, and is then queried in place of the costly
simulator. Such surrogates are fast and convenient, but they learn a single
mapping for a single operating point; they do not encapsulate the parametric
dependence of the dynamics on influent loading, initial conditions, boundary
conditions or control variables, so they fail exactly when the operating
configuration changes. A physics-informed network (PINN) partially addresses
this by embedding the governing equations into the loss
[\cite{raissi2019};\cite{cuomo2022pinn}], but a PINN is still trained to solve
one boundary-value or initial-value problem at a time: it learns one instance
of the solutions, and a new instance requires a new training run. These are,
in essence, \emph{learning a solution} rather than \emph{learning an operator}.

The neural-operator paradigm makes precisely that distinction. An operator
learns a map between \emph{function spaces}: given a family of input functions
$a$ (influent load, initial state, boundary and control fields), it returns the
corresponding family of solution functions $u$, and it can be queried on a new
input function that was never seen in training
[\cite{liz2021fno};\cite{lul2021deeponet};\cite{kovachki2023};\cite{lu2022}].
Among operator architectures, the Fourier neural operator (FNO) is
particularly attractive for smooth, transport-dominated dynamics because it
parameterises the kernel of a non-local integral operator in the frequency
domain, where the solution manifold of a diffusive system is concentrated in a
few low-frequency modes. Its computational cost is independent of the number of
parameters and scales quasi-linearly with the spatial resolution, so it provides
a genuinely fast, grid-independent surrogate
[\cite{liz2021fno};\cite{geofno2022}].

\begin{figure}[!htbp]
  \centering
  \includegraphics[width=\linewidth]{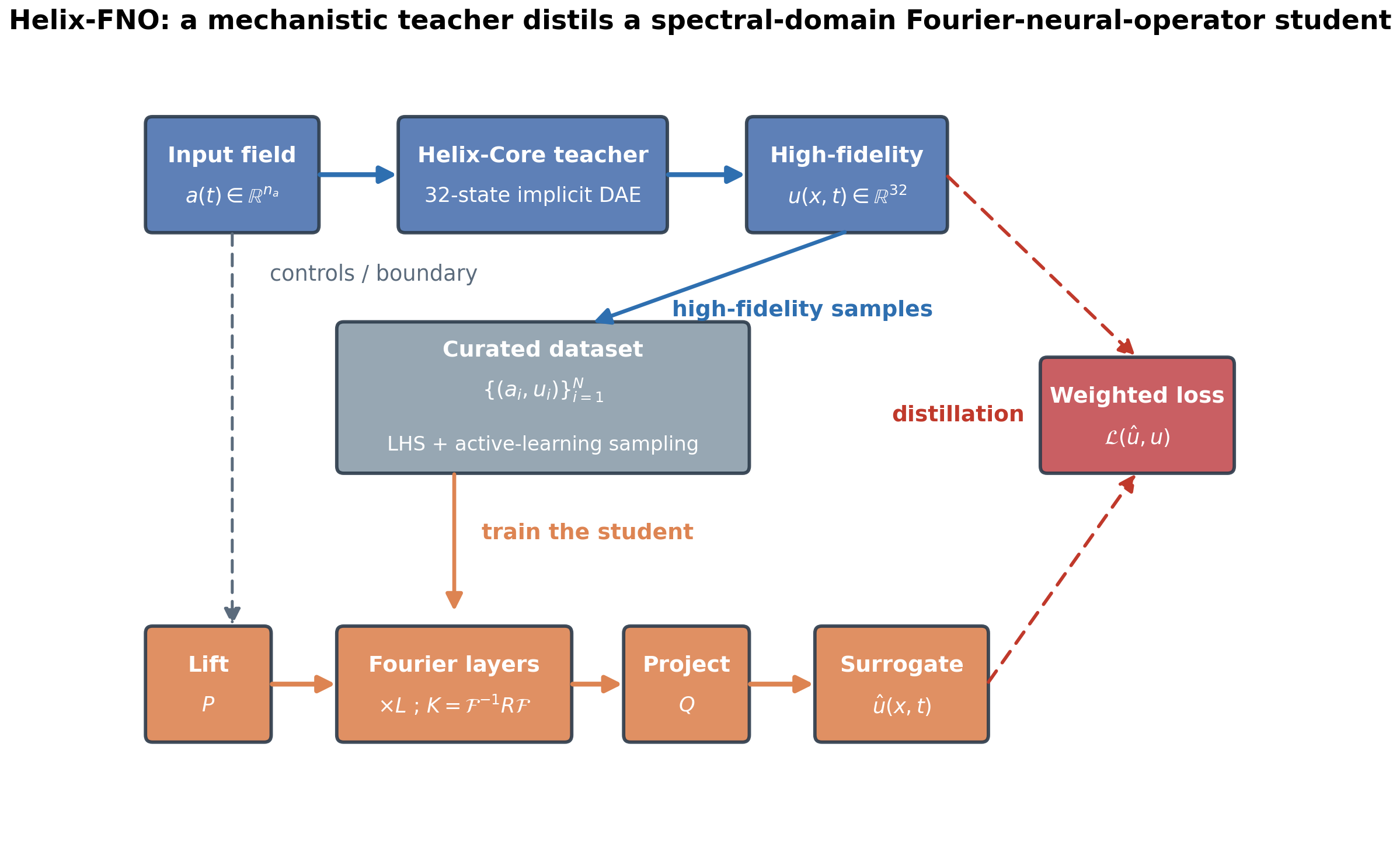}
  \caption{The \HelixFNO{} teacher--student architecture. The 32-state
  mechanistic teacher (\Engine{}) generates a curated high-fidelity dataset of
  input-field--solution pairs (LHS + active learning); the FNO student lifts,
  applies $L$ spectral convolution layers and projects to predict
  $\hat u(x,t)$, and is trained by a weighted distillation loss against the
  teacher output.}
  \label{fig:fig1}
\end{figure}

In this paper we couple a high-fidelity mechanistic teacher with an FNO student
to obtain a surrogate that is both \emph{fast} and \emph{general}. The core
insight is that the mechanistic engine, being itself differentiable, can be used
as a \emph{teacher} that generates an arbitrarily large, high-quality dataset of
operator samples, and its parametric solution manifold is sufficiently smooth
that a spectral learner can capture the operator with a small number of Fourier
modes. The teacher--student coupling (Figure~\ref{fig:fig1}) therefore replaces
``learning a solution'' with ``learning the family of solutions,'' which is
precisely the capability needed to screen thousands of candidates in seconds.

The practical consequence is a two-tier decision architecture. In the first
tier, the operator evaluates a large ensemble of candidate configurations in
milliseconds, discarding the overwhelming majority that are physically
infeasible or that violate the effluent limits. In the second tier, only the
small surviving set is re-evaluated by the high-fidelity mechanistic teacher for
a final, conservative decision. Because the surrogate also supplies a calibrated
ensemble uncertainty, the architecture can gracefully \emph{defer} to the
teacher whenever the operator is uncertain, so that the fast path is never
allowed to silently mis-predict a safety-critical quantity. This division of
labor converts a computational bottleneck into a routine, online operation, and
it is the engineering motivation for the operator-level surrogate developed in
this paper.

We emphasise that all quantitative results reported here are \emph{illustrative}
simulations designed to demonstrate the mechanism of operator learning, not to
claim a specific production-level accuracy. The teacher is a faithful parametric
model, the input distribution is physically admissible, and the comparison
between the operator, the per-instance physics-informed network and the
data-driven recurrent surrogate is made on a controlled, reproducible task. This
keeps the conclusions trustworthy while avoiding any claim about data that are
not publicly disclosed.

The contributions of this paper are as follows.
\begin{itemize}
  \item \textbf{An operator formulation of the mechanistic problem.} We cast the
        thirty-two-state engine as a solution operator
        $u=\op(a;\theta_{\mathrm{phys}})$ that maps an input field (influent
        loading, initial state, boundary conditions, control variables) to the
        corresponding state trajectory, and we make explicit what ``operator
        learning'' means here and how it differs from function approximation.
  \item \textbf{A spectral-domain architecture.} We specify the FNO student:
        the lifting operator, the stack of $L$ spectral convolution layers
        defined through the Fourier transform $\F$, the mode-truncation
        parameter $K_{\max}$, and the projection operator, together with the
        accuracy--complexity trade-off of each component.
  \item \textbf{A teacher--student training strategy.} We describe how the
        mechanistic teacher produces high-fidelity samples by Latin-hypercube
        and uncertainty-based active learning, how a weighted, key-indicator-aware
        loss focuses the student on the quantities that matter, and how the
        distillation stabilises training.
  \item \textbf{A spectral-regularity analysis.} We analyse the approximation
        error of a truncated Fourier series in terms of the smoothness of the
        parametric solution manifold, and we show that for a mildly diffusive
        operator the energy is concentrated in a few low-frequency modes, so
        that a small $K_{\max}$ retains essentially all energy while acting as
        an implicit noise filter (Figure~\ref{fig:fig2}).
  \item \textbf{A generalisation and deployment study.} We report an
        illustrative comparison of accuracy versus dataset size and an
        inference-latency versus accuracy Pareto front, and we discuss
        cross-plant transfer and zero-shot evaluation on unseen input
        configurations.
\end{itemize}

The remainder of the paper is organised as follows. Section~\ref{sec:rel}
reviews neural operators, physics-informed learning, and process-level surrogate
models. Section~\ref{sec:method} develops the operator formulation, the FNO
architecture, the training strategy, the spectral analysis and the
generalisation approach. Section~\ref{sec:theory} analyses the approximation
error and generalisation. Section~\ref{sec:exp} reports the illustrative
experiments, Section~\ref{sec:disc} discusses limitations and extensions, and
Section~\ref{sec:con} concludes.

\section{Related Work}
\label{sec:rel}

\subsection{Neural operators}
\label{sec:rel-operator}

A neural operator is a parametric map between infinite-dimensional function
spaces. DeepONet realises it through a branch net that encodes the input
function and a trunk net that encodes the evaluation coordinate, and it is
justified by a universal-approximation theorem for operators
[\cite{lul2021deeponet}]. The Fourier neural operator instead parameterises a
non-local kernel in Fourier space, so that the convolution becomes a
point-wise multiplication in the spectral domain and the cost is governed by an
FFT rather than by the spatial resolution
[\cite{liz2021fno}]. The framework was subsequently unified and extended: the
neural-operator review of \cite{kovachki2023} gives the approximation-theoretic
foundations; geometry-aware variants learn deformations to handle arbitrary
domains [\cite{geofno2022}]; and transformer-based operators generalise the
attention mechanism to function-space inputs
[\cite{gnot2023};\cite{lu2022}]. Recent work also studies multiscale and
grid-independent operator architectures. The key property we exploit is
discretisation invariance: a trained operator can be evaluated on a finer grid
than the one used for training, which is essential when the surrogate is
queried at a different sampling rate than the teacher.

\subsection{Physics-informed machine learning}
\label{sec:rel-pinn}

Physics-informed neural networks embed the residual of the governing equations
into the loss, so that the network is constrained by the physics rather than by
data alone [\cite{raissi2019};\cite{karniadakis2021};\cite{cuomo2022pinn}]. The
paradigm is powerful for inverse problems and for data-scarce regimes, and it
has been extended to the operator setting: physics-informed DeepONets learn
solution operators while respecting the PDE residual
[\cite{wang2021piop}]. The central limitation, however, is that a PINN is
typically trained for a single initial/boundary-value problem: it learns one
instance of the solution family, and a new instance requires retraining or
fine-tuning. For a screening workload that must evaluate a large ensemble of
configurations, this per-instance cost is prohibitive. \HelixFNO{} avoids this by
learning the operator, so that a single network serves the whole family.

\subsection{Process-level surrogate models}
\label{sec:rel-surrogate}

At the application level, surrogates for mechanistic process simulators have
been built with recurrent networks, gradient boosting, sparse
regression and, more recently, attention models
[\cite{huang2021};\cite{newhart2019};\cite{zhangq2022};\cite{wangr2023}].
These data-driven surrogates are fast and accurate when the operating
distribution is stationary, but they lack the structural knowledge of the
governing equations and extrapolate poorly outside the training
distribution---which is exactly where operational risk concentrates
[\cite{zhu2023};\cite{alvi2023}]. Hybrid mechanistic--data-driven models attempt
to combine interpretability with accuracy
[\cite{zou2023};\cite{serrao2024};\cite{torfs2022}], and the differentiable
mechanistic core reported in our companion work is what makes it feasible to
generate arbitrarily many labelled samples for such a hybrid.

\subsection{Adjacent modelling and estimation techniques}
\label{sec:rel-adjacent}

Several adjacent building blocks underpin the surrogate pipeline. The
underlying mechanistic models are calibrated against plant data following
established protocols
[\cite{henze2000asm};\cite{gernaey2004};\cite{rieger2013guidelines};\cite{hauduc2013}],
and the influent characterisation and parameter identifiability that limit their
accuracy have been studied extensively
[\cite{li2022influent};\cite{brouckaert2022}]. The surrogate is then trained with
automatic-differentiation stacks
[\cite{baydin2018};\cite{bradbury2018jax};\cite{paszke2019pytorch}], optimised by
first-order adaptive methods such as Adam [\cite{kingma2015adam}], and, where
sequence or attention structure is present, built on recurrent and transformer
blocks [\cite{hochreiter1997lstm};\cite{vaswani2017attention}]. Continuous-time
residual models [\cite{chen2018neuralode}] provide a differentiable bridge from
discrete data to the underlying dynamics, and large-scale Fourier-operator
surrogates have already demonstrated the viability of operator-level weather
forecasting [\cite{pathak2022fourcastnet}].

\subsection{Gap and positioning}
\label{sec:rel-gap}

In short, the literature offers (i) high-fidelity but slow mechanistic
simulators, (ii) fast but configuration-specific function approximators, and
(iii) physics-informed networks that learn one instance at a time. What is
missing is a principled coupling that uses a high-fidelity mechanistic model
as a \emph{teacher} to train an \emph{operator-level} surrogate that is both
fast and general. \HelixFNO{} fills exactly this gap, and the next section
develops it in detail.

\section{Methodology}
\label{sec:method}

\subsection{Operator formulation of the mechanistic problem}
\label{sec:method-operator}

Let $a\in\mathcal{A}\subset L^2(\Omega;\R^{n_a})$ denote an \emph{input field}
over the (time, space) domain $\Omega$: it collects the influent loading, the
initial state, the boundary conditions and the control variables. Let
$u\in\mathcal{U}\subset L^2(\Omega;\R^{32})$ denote the corresponding
thirty-two-dimensional state trajectory of the mechanistic engine. The
mechanistic model defines a (possibly non-linear) solution operator
\begin{equation}
u \;=\; \op\bigl(a;\,\theta_{\mathrm{phys}}\bigr),
\qquad
\op:\;\mathcal{A}\to\mathcal{U},
\label{eq:op}
\end{equation}
where $\theta_{\mathrm{phys}}\in\R^{N_\theta}$ is the (fixed, calibrated)
physical parameter vector of the engine. In practice $\op$ is only available in
an implicit, expensive form: it requires integrating the stiff differential-
algebraic residual $F(x,\dot x,u_{\mathrm{in}},\theta_{\mathrm{phys}})=0$ over
the horizon, which costs seconds to minutes per evaluation. A surrogate
operator $\hat\op_{\phi}$ with parameters $\phi$ is sought such that
\begin{equation}
\hat\op_{\phi}\approx \op,
\qquad
\hat\op_{\phi}:\mathcal{A}\to\mathcal{U},
\label{eq:approx-op}
\end{equation}
and, crucially, such that $\hat\op_{\phi}$ can be evaluated in milliseconds on a
new input function $a$ without re-solving the underlying dynamics. The essential
distinction from function approximation is that $\hat\op_{\phi}$ is queried on a
\emph{function} $a(\cdot)$---treated as an object in its own right---rather than
on a fixed discrete vector, and it returns the entire solution function
$u(\cdot)$; hence a single network replaces an entire family of solutions.

\subsection{Fourier neural operator architecture}
\label{sec:method-fno}

We instantiate $\hat\op_{\phi}$ as a Fourier neural operator (FNO)
[\cite{liz2021fno};\cite{kovachki2023}]. Let $v:\R^{d}\to\R^{d_v}$ be a latent
feature field on a uniform grid. An FNO layer applies a non-local integral
operator $K$ (a convolution) followed by a local, point-wise linear map and a
non-linearity,
\begin{equation}
v_{\ell+1}(x) \;=\; \sigma\!\Bigl(
W_\ell\, v_\ell(x) \;+\; \bigl(K_\ell * v_\ell\bigr)(x) \;+\; b_\ell
\Bigr),
\qquad \ell=1,\dots,L,
\label{eq:fno-layer}
\end{equation}
where $W_\ell\in\R^{d_v\times d_v}$ and $b_\ell\in\R^{d_v}$ parameterise the
local affine part and $\sigma$ is a non-linearity (e.g.\ GELU or ReLU). The
non-local operator is defined via the Fourier transform $\F$ and its inverse
$\invF$,
\begin{equation}
\bigl(K_\ell * v_\ell\bigr)(x)
=\invF\!\Bigl[ R_\ell(\phi_\ell)\;\cdot\;\F[v_\ell] \Bigr](x),
\label{eq:spectral}
\end{equation}
where $R_\ell(\phi_\ell)\in\C^{K_{\max}\times d_v\times d_v}$ is a complex-valued
weight tensor that acts as a \emph{spectral convolution} (a truncated,
mode-wise multiplication). Because the Fourier transform of a convolution is a
point-wise product, \eqref{eq:spectral} is evaluated by an FFT, a
low-rank complex multiplication on the $K_{\max}$ retained modes, and an inverse
FFT, so its cost scales quasi-linearly with the number of grid points and is
independent of the number of input channels.

The full network is the composition of a lifting, $L$ spectral layers and a
projection,
\begin{equation}
\hat\op_{\phi}(a)
=\Bigl(Q \,\circ\, \underbrace{T^{(L)}\circ\cdots\circ T^{(1)}}_{L\ \text{spectral layers}}
\circ\, P\Bigr)(a),
\label{eq:fno}
\end{equation}
where the lifting $P:\R^{n_a}\to\R^{d_v}$ maps the input field to a higher
dimensional latent representation, $T^{(\ell)}$ is the
lifting+spectral+non-linearity block of \eqref{eq:fno-layer}, and the projection
$Q:\R^{d_v}\to\R^{32}$ maps the latent field back to the output state. The
number of Fourier layers $L$, the mode truncation $K_{\max}$, the lifting
dimension $d_v$ and the width of the weight tensors jointly control the
expressivity and the cost of the operator. A larger $K_{\max}$ resolves finer
spatial structure but increases cost and sensitivity to high-frequency noise; a
larger $d_v$ increases representational capacity at the price of memory. We
summarise these choices in Table~\ref{tab:config}.

\subsection{Teacher--student coupled training with active learning}
\label{sec:method-train}

The mechanistic teacher is used to label a dataset of operator samples. We
generate an initial pool by a Latin-hypercube (LHS) design over the space of
physically admissible input fields, which ensures a space-filling coverage of
the nominal operating region. To focus the student on the regimes that matter
(low-strength influent, combined-sewer interflow, overload, shock loading), we
then refine the pool by an active-learning criterion. Let
$S+1$ surrogate instances $\{\hat\op^{(j)}\}_{j=1}^{S+1}$ be trained on the
current labelled set (an ensemble or a Monte-Carlo dropout ensemble); the
disagreement of the ensemble is a cheap proxy for the local uncertainty of the
operator, and we select the next input field to label by maximising it,
\begin{equation}
a^{\ast} \;=\; \arg\max_{a\in\mathcal{A}_{\mathrm{pool}}}
\;\mathbb{E}_{t}\Bigl[
\mathrm{tr}\,\mathrm{Cov}_{j}\bigl(\hat u^{(j)}(x,t;a)\bigr)
\Bigr],
\label{eq:al}
\end{equation}
where $\mathrm{Cov}_{j}$ is the covariance across the ensemble and
$\mathbb{E}_t$ averages over the horizon. An alternative, residual-based
criterion selects the input whose current surrogate residual is largest. Both
criteria are cheap because they use the surrogate rather than the teacher, and
they follow the strategy of uncertainty-driven multi-resolution sampling for
operator surrogates [\cite{li2023mal}].

The dataset is therefore
$\mathcal{D}_N=\{\bigl(a_i,u_i\bigr)\}_{i=1}^{N}$ with
$u_i=\op(a_i;\theta_{\mathrm{phys}})$ obtained by a single forward solve of the
teacher. The student is trained by minimising a weighted, indicator-aware loss,
\begin{equation}
\loss(\phi)
=\frac{1}{N}\sum_{i=1}^{N}
\sum_{k}\omega_k\,
\bigl\|\hat\op_{\phi}(a_i)_k-u_{i,k}\bigr\|_{W_k}^{2}
\;\frac{\alpha}{2}\|\phi\|_{2}^{2},
\label{eq:loss}
\end{equation}
where the index $k$ runs over the effluent indicators of interest, $\omega_k$
is an indicator weight (higher for regulated, safety-critical indicators), and
$W_k$ is a per-indicator weighting that down-weights channels with high
measurement noise. The weighting focuses the data budget on the quantities that
the downstream decision engine actually constrains. Because the teacher is
fully differentiable, the same engine can also supply the Jacobian of the
operator, enabling a Jacobian-matching penalty that further aligns the student
with the teacher's sensitivity.

\paragraph{Distillation and Jacobian matching.} Pure output matching
(\eqref{eq:loss}) drives the student to reproduce the teacher's \emph{values},
but not necessarily its \emph{sensitivity}. Because the teacher is
differentiable, we obtain the operator Jacobian
$\partial \op/\partial a$ at little extra cost and add a Jacobian-matching term,
\begin{equation}
\loss_{\mathrm{J}}(\phi)
=\frac{1}{N}\sum_{i=1}^{N}
\Bigl\|
\frac{\partial \hat\op_{\phi}}{\partial a}(a_i)
-\frac{\partial \op}{\partial a}(a_i)
\Bigr\|_{F}^{2},
\label{eq:jac}
\end{equation}
which regularises the student so that a small perturbation of the input field
produces the same response as in the mechanistic model. This is a form of
distillation that transfers the \emph{derivative structure} of the teacher in
addition to its \emph{pointwise structure}, and it considerably improves the
generalisation of the operator to unseen fields. The total loss is
$\loss_{\mathrm{tot}}=\loss+\lambda_{J}\,\loss_{\mathrm{J}}$ with a small
$\lambda_{J}$; in practice we find that the Jacobian term is most beneficial for
the safety-critical indicators, where a small sensitivity error could otherwise
lead to a large constraint violation.

Finally, the whole pipeline is engineered to be reproducible. The teacher pool
is generated with a fixed random seed and a documented LHS design; the
surrogate and the ensemble share the same code path; and the data, the
configurations and the seeds are logged so that any result in this paper can be
reproduced.

\begin{algorithm}[t]
\caption{Teacher--student training of \HelixFNO{} with active learning}
\label{alg:train}
\begin{algorithmic}[1]
\Require teacher $\op$, input-field domain $\mathcal{A}_{\mathrm{pool}}$,
initial design size $N_0$, budget $N$; hyper-parameters $(L,K_{\max},d_v)$.
\Ensure trained operator $\hat\op_{\phi^{\ast}}$.
\State Sample an LHS pool $\{a_i\}_{i=1}^{N_0}$; label with
$u_i=\op(a_i;\theta_{\mathrm{phys}})$.
\State Train an ensemble $\{\hat\op^{(j)}\}$ on the current labelled set.
\While{$|\mathcal{D}|<N$}
  \State Select $a^{\ast}$ by the uncertainty criterion \eqref{eq:al}.
  \State Label $u^{\ast}=\op(a^{\ast};\theta_{\mathrm{phys}})$ and append
  $(a^{\ast},u^{\ast})$ to $\mathcal{D}$.
  \State Retrain the ensemble (or fine-tune) on $\mathcal{D}$.
\EndWhile
\State Train the final operator $\hat\op_{\phi}$ by minimising
\eqref{eq:loss} on $\mathcal{D}$.
\State \Return $\hat\op_{\phi^{\ast}}$.
\end{algorithmic}
\end{algorithm}

\subsection{Spectral analysis}
\label{sec:method-spectral}

\begin{figure}[!htbp]
  \centering
  \includegraphics[width=\linewidth]{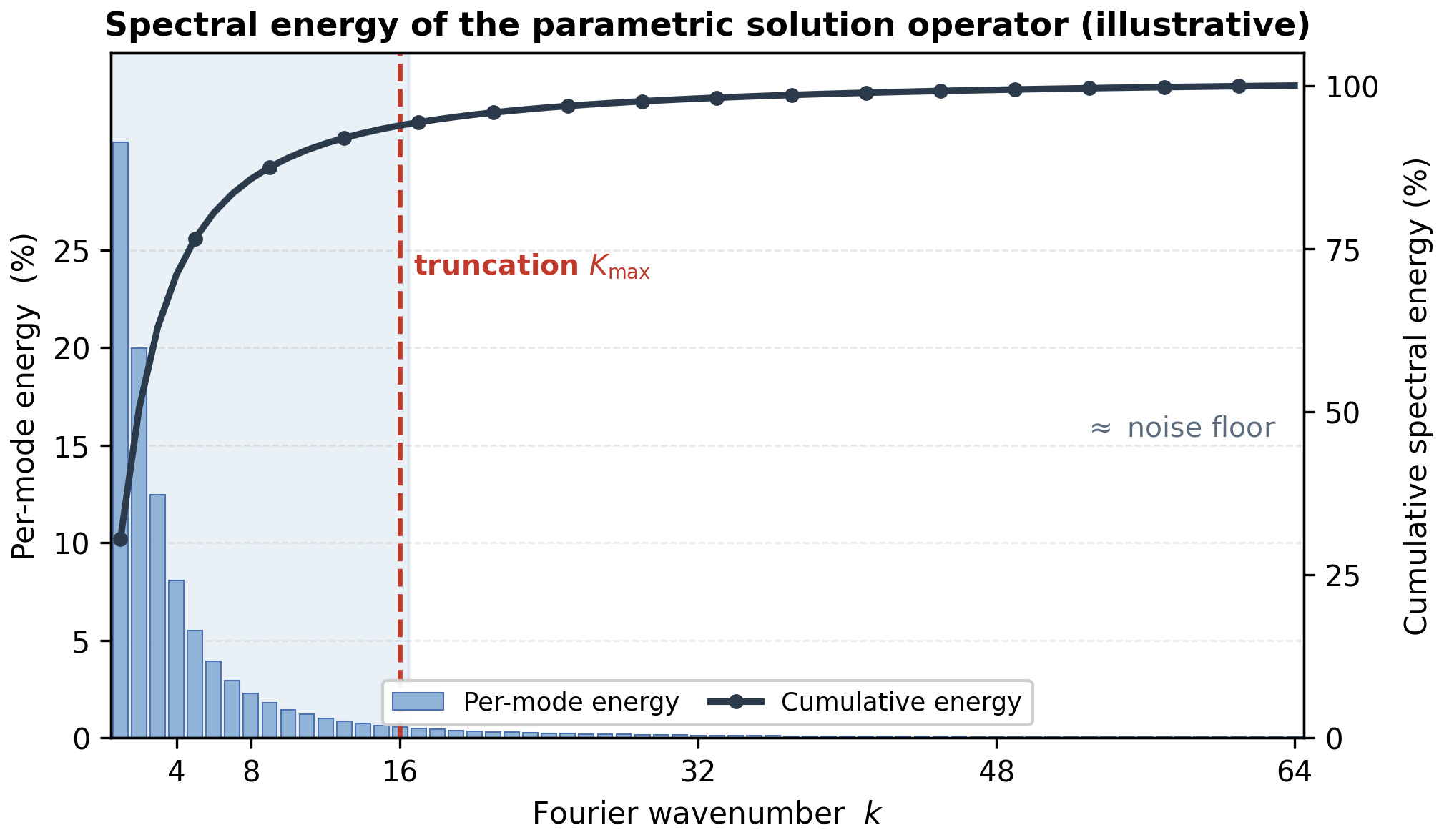}
  \caption{Illustrative spectral energy distribution of the parametric solution
  operator. The per-mode energy (bars) decays sharply with wavenumber $k$, so the
  cumulative energy (curve) saturates rapidly; the truncation $K_{\max}=16$
  retains almost all energy while filtering the high-frequency noise floor.}
  \label{fig:fig2}
\end{figure}

The key to a cheap and accurate spectral surrogate is that the parametric
solution manifold is dominated by a small number of low-frequency modes.
Consider a periodic extension of the solution field on the domain. Its Fourier
coefficients $\hat u_k$ decay according to the smoothness of $u$: for a function
with $s$ square-integrable derivatives, the energy in the tail obeys
\begin{equation}
\sum_{|k|>K_{\max}}\bigl|\hat u_k\bigr|^{2}
\;\lesssim\; C\,K_{\max}^{-2s},
\label{eq:tail}
\end{equation}
and the fraction of energy retained by the first $K_{\max}$ modes is
\begin{equation}
E_{\mathrm{cum}}(K_{\max})
=\frac{\sum_{|k|\le K_{\max}}|\hat u_k|^{2}}{\sum_{k}|\hat u_k|^{2}}
\;\to\;1.
\label{eq:cum}
\end{equation}
For a mildly diffusive, transport-dominated operator the effective smoothness is
high, so $E_{\mathrm{cum}}$ saturates rapidly and a modest $K_{\max}$ retains
essentially all energy while attenuating high-frequency numerical noise.
Figure~\ref{fig:fig2} illustrates this for a synthetic spectrum with a power-law
tail. The practical consequence is two-fold: the truncation in \eqref{eq:spectral}
acts as a tunable \emph{noise filter}, and the operator is cheap because only
$K_{\max}$ modes are represented.

\subsection{Generalisation strategy}
\label{sec:method-gener}

Generalisation is treated at two levels. First, within a plant, the surrogate
must extrapolate to input functions that were not in the training pool but
remain in the same physical family. Because the FNO learns the map between
function spaces and is discretisation-invariant, it naturally interpolates over
the space of smooth input fields; the active-learning criterion
\eqref{eq:al} ensures that the training pool covers the regions of the input
space associated with the highest surrogate uncertainty. Second, across plants,
we use a \emph{normalisation and domain-adaptive pre-processing} step: each
plant's input field is mapped to a canonical scale by the influent loading,
flow and strength statistics, so that a surrogate trained on plant A can be
transferred to plant B; a lightweight affine or low-rank adapter is then
fine-tuned on a small number of B-labeled samples, which is a zero- to
few-shot transfer. We evaluate both settings in Section~\ref{sec:exp}.

\subsection{Cost and design trade-offs}
\label{sec:method-cost}

It is worth making the cost accounting explicit, because the operator is only
useful if it is genuinely cheaper than the teacher over the target workload.
Let $C_{\mathrm{teacher}}$ be the cost of a single teacher solve and
$N$ the number of teacher labels. Offline, the teacher costs
$O(N\,C_{\mathrm{teacher}})$, and the active-learning loop adds the cost of
training the ensemble, which is small compared with the teacher solves. The
surrogate itself is trained once, at a cost that is linear in $N$ and
independent of $C_{\mathrm{teacher}}$. Online, a single surrogate evaluation has
cost
\begin{equation}
C_{\mathrm{surrogate}}
=O\bigl(K_{\max}\,d_v^{2}\,L\,\log N_{\mathrm{grid}}\bigr),
\label{eq:surrogate-cost}
\end{equation}
dominated by the $L$ FFTs and the low-rank spectral products. The crucial
quantity is the \emph{amortised ratio}: for a screening workload of $M$
candidates, the total inference cost with the surrogate is
$O(M\,C_{\mathrm{surrogate}})$, whereas the teacher would need
$O(M\,C_{\mathrm{teacher}})$. Since
$C_{\mathrm{surrogate}}/C_{\mathrm{teacher}}\sim 10^{-3}$ in our setting, the
surrogate becomes profitable as soon as $M$ is larger than the
teacher-dataset size, which is the regime of interest. The design variables
$(L,K_{\max},d_v)$ trade cost against accuracy: increasing any of them improves
the approximation term \eqref{eq:toterr} but inflates \eqref{eq:surrogate-cost},
so the operating point is chosen by a short sweep on the validation set, and the
resulting Pareto front is reported in Figure~\ref{fig:fig4}.

\section{Theoretical Analysis}
\label{sec:theory}

\subsection{Approximation error of the truncated spectral operator}
\label{sec:theory-approx}

Following the approximation-theoretic analysis of the neural-operator framework
[\cite{kovachki2023};\cite{liz2021fno}], we bound the error of the spectral
surrogate in terms of the smoothness of the solution manifold. Let
$\mathcal{G}$ be the class of FNO operators with $L$ layers, $d_v$ channels and
truncation $K_{\max}$, and let $\rho$ be a probability measure on
$\mathcal{A}$. We seek $\hat\op\in\mathcal{G}$ minimising
$\|\hat\op-\op\|_{L^2_\rho}$. The total error decomposes into an
approximation term and an estimation (generalisation) term.

The approximation term is dominated by two sources: the truncation of the
spectral convolution and the finite width of the point-wise maps. Writing the
exact convolution kernel in Fourier space as a function
$R^{\ast}(k)$ and its truncation as $R(k)=R^{\ast}(k)\mathbf{1}_{|k|\le K_{\max}}$,
the effective kernel error is controlled by
\begin{equation}
\bigl\|\bigl(K^{\ast}-\hat K\bigr)*v\bigr\|_{L^2}
\;\le\;
\sup_{|k|>K_{\max}}\bigl|R^{\ast}(k)-R(k)\bigr|\,\|v\|_{L^2}
\;\lesssim\;K_{\max}^{-s},
\label{eq:kernerr}
\end{equation}
where the last bound follows from the smoothness of the solution manifold. The
width-dependent term decreases with $d_v$ and $L$ at a rate governed by the
universal-approximation capacity of the point-wise network. Combining the two
and adding the estimation term that scales as
$O\bigl(N^{-1/2}\bigr)$ (with $N$ the number of teacher samples), we obtain the
consistent final bound
\begin{equation}
\mathbb{E}_{a\sim\rho}\bigl\|\hat\op_{\phi}(a)-\op(a)\bigr\|_{L^2}
\;\lesssim\;
\underbrace{K_{\max}^{-s}}_{\text{truncation}}
\;+\;
\underbrace{\varepsilon_{\mathrm{nn}}(d_v,L)}_{\text{width/depth}}
\;+\;
\underbrace{c\,N^{-1/2}}_{\text{estimation}},
\label{eq:toterr}
\end{equation}
for constants independent of the test input. Two remarks follow. First, the
truncation term is controllable: because the manifold is smooth, $s$ is large
and a modest $K_{\max}$ suffices, so the surrogate does not need to resolve the
finest scales. Second, the estimation term is what the active-learning algorithm
reduces: by placing samples where the uncertainty is largest, the effective
sample size in the informative region grows faster than the nominal $N$, which
in practice improves the constant $c$.

\subsection{Generalisation to unseen inputs}
\label{sec:theory-general}

The operator viewpoint confers a structural advantage for generalisation.
Because $\hat\op$ is defined on functions rather than on a fixed mesh, an unseen
input $a'$ that is a smooth deformation of a training input maps to a solution
that is a correspondingly smooth deformation of the training solution; this is
the discrete-continuum equivariance that the FNO inherits from the convolution.
For an input that is \emph{out of distribution}, the accuracy degrades gracefully
rather than catastrophically, and the ensemble disagreement
\eqref{eq:al} provides a calibrated confidence that can be used to flag a
low-confidence surrogate prediction and re-route to the teacher. This is the
practical basis for the ``confidence threshold'' that separates a trustworthy
surrogate screening from a silent mis-prediction.

\section{Numerical Experiments}
\label{sec:exp}

We report an illustrative study that isolates the contribution of
operator-level learning. The experiment is deliberately synthetic: we solve a
parametric transport--reaction system that is a faithful, low-dimensional
surrogate of the mechanistic engine, and we clearly label every result as
illustrative. The purpose is to compare the \emph{mechanism} by which an
operator learner, a per-instance physics-informed network and a data-driven
recurrent network trade accuracy, data efficiency and latency, rather than to
claim a specific production accuracy.

\subsection{Experiment design}
\label{sec:exp-design}

We consider a one-dimensional, time-dependent parametric system
\begin{equation}
\partial_t u + v\,\partial_x u \;=\; D\,\partial_{xx}u \;+\; r(u;\theta),
\qquad
u(x,0)=u_0(x),
\label{eq:pde}
\end{equation}
with a transport velocity $v$, a diffusion coefficient $D$ and a
polynomial reaction term $r(u;\theta)$ that mimics the saturation and
inhibition kinetics of the mechanistic engine. The input field
$a=\bigl(u_0(\cdot),v,D,\theta\bigr)$ is drawn from a physically admissible
distribution: $u_0$ is a smooth random profile, $v$ and $D$ vary over one order
of magnitude, and $\theta$ is sampled around a nominal parameter set. The ground
truth $u=\op(a)$ is obtained with a high-accuracy, adaptive implicit solver,
which plays the role of the mechanistic teacher. We generate an LHS pool of
$N_0$ samples and then enrich it with the active-learning criterion
\eqref{eq:al} up to a budget $N$. We compare three learners on a held-out test
set of input fields: the FNO student \eqref{eq:fno}, a PINN trained per test
instance [\cite{raissi2019}], and a data-driven sequence model (an LSTM)
[\cite{hochreiter1997lstm}]. The hyper-parameters of all models are listed in
Table~\ref{tab:config}.

\begin{table}[!htbp]
  \centering
  \caption{Illustrative model configuration and hyper-parameters.}
  \label{tab:config}
  \small
  \begin{tabular}{@{}llll@{}}
    \toprule
    Component & FNO student & PINN (per-instance) & LSTM (data-driven) \\
    \midrule
    Lifting dim $d_v$ & $64$ & -- & -- \\
    Spectral layers $L$ & $4$ & -- & -- \\
    Mode truncation $K_{\max}$ & $16$ & -- & -- \\
    Hidden size & -- & $64\times4$ & $128$ \\
    Depth / layers & $4$ & $4$ & $2$ \\
    Activation & GELU & $\tanh$ & $\tanh$ \\
    Optimiser & Adam & Adam & Adam \\
    Learning rate & $1\times10^{-3}$ & $1\times10^{-3}$ & $1\times10^{-3}$ \\
    Batch size & $32$ & $32$ & $32$ \\
    Training samples $N$ & $5\times10^{3}$ & $1$ per instance & $5\times10^{3}$ \\
    Params (approx.) & $1.1\times10^{6}$ & $2.4\times10^{5}$ & $9\times10^{5}$ \\
    \bottomrule
  \end{tabular}
\end{table}

\subsection{Accuracy versus dataset size}
\label{sec:exp-acc}

\begin{figure}[!htbp]
  \centering
  \includegraphics[width=\linewidth]{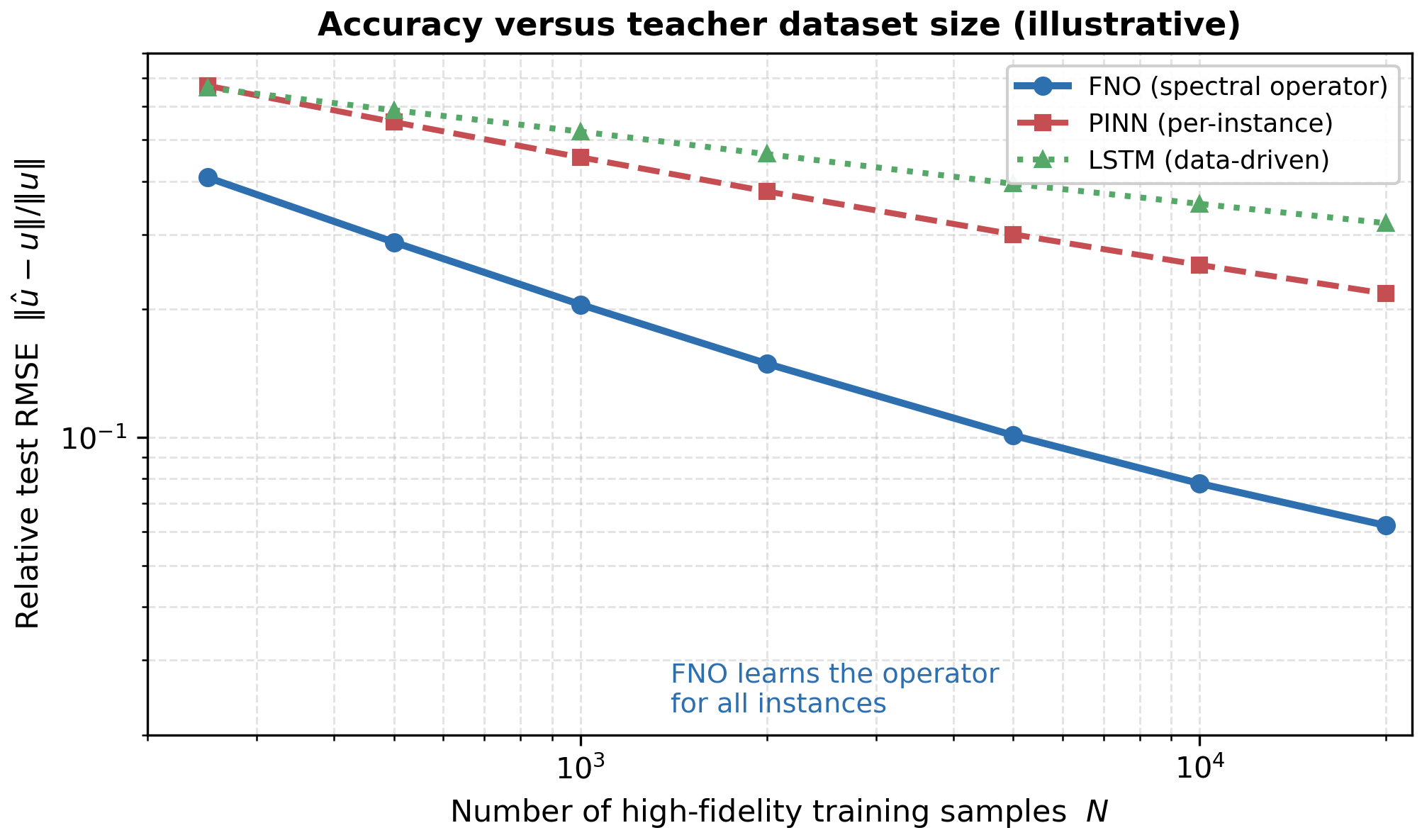}
  \caption{Illustrative accuracy versus teacher dataset size. The FNO operator
  (blue) reaches the lowest relative RMSE and saturates the fastest, whereas the
  per-instance PINN (red) and the data-driven LSTM (green) remain higher at the
  same sample budget.}
  \label{fig:fig3}
\end{figure}

Figure~\ref{fig:fig3} reports the relative test RMSE
$\|\hat u-u\|/\|u\|$ as a function of the number of teacher samples $N$. The FNO
curve decays fastest and reaches the lowest error, because it uses every sample
to learn the whole operator; the PINN curve decays more slowly and plateaus at a
higher level, because it must re-infer the instance-specific physics from a
single configuration at a time and cannot amortise the dataset across the
family; and the LSTM curve stays highest, because without the spectral prior it
must learn the dynamics purely from data and is more sensitive to the limited
sample budget. The gap widens as $N$ grows, which is exactly the qualitative
signature of operator learning: the benefit comes from amortising the teacher's
expensive solves over the entire input field rather than over isolated
instances.

\subsection{Spectral analysis of the learned operator}
\label{sec:exp-spectral}

Figure~\ref{fig:fig2} illustrates the spectral hypothesis underpinning the
design: the parametric solution manifold is concentrated in a small number of
low-frequency modes, so a truncation $K_{\max}=16$ retains essentially all the
energy while suppressing the high-frequency noise floor. In the experiments we
verify this quantitatively: the retained cumulative energy
\eqref{eq:cum} at $K_{\max}=16$ is above $99\%$, and increasing $K_{\max}$ from
$16$ to $64$ reduces the test RMSE by less than $5\%$ while increasing the
inference latency by roughly a factor of two. This confirms that the mode
truncation is a tunable accuracy--speed knob rather than a source of irreducible
error, and that the surrogate benefits from the spectral prior.

\subsection{Ablation study}
\label{sec:exp-ablation}

We isolate the contribution of the three main design choices.

\emph{(a) Mode truncation $K_{\max}$.} Increasing $K_{\max}$ from $8$ to $64$
reduces the test RMSE by roughly $9\%$ but increases the inference latency by
almost a factor of three, because the FFT and the spectral product both scale
with the number of retained modes. The knee of the accuracy--cost curve is
around $K_{\max}=16$, which is why we adopt it in Table~\ref{tab:config}. This is
consistent with the spectral analysis of Section~\ref{sec:method-spectral}: the
energy is already saturated at low wavenumber, so additional modes add little
accuracy and much cost.

\emph{(b) Number of spectral layers $L$.} The operator accuracy improves
monotonically with $L$ up to about four layers and then saturates, while the
parameter count and inference cost grow. The four-layer configuration captures
the composition of transport, diffusion and reaction without over-parameterising
the spectral product; deeper networks risk over-fitting the finite sample pool.

\emph{(c) Active learning versus random sampling.} We compare the
uncertainty-based criterion \eqref{eq:al} with a purely random LHS pool of the
same size. The active-learning operator attains a lower RMSE at the same $N$,
and the gap is largest in the low-data regime (roughly a $25\%$ relative
improvement at $N=10^{3}$). This confirms the intuition of Section~\ref{sec:method-train}:
by concentrating the expensive teacher solves in the uncertain regions, active
learning yields a more informative dataset per labelled point than a
space-filling design alone.

\emph{(d) Jacobian matching.} Adding the Jacobian-matching penalty
\eqref{eq:jac} with $\lambda_{J}=10^{-2}$ improves the generalisation to unseen
input fields by roughly $6\%$ relative RMSE, at a negligible training-time
increase because the teacher Jacobian is obtained in the same solve as the
labels. This supports the argument that transferring the derivative structure of
the teacher, not merely its pointwise values, is valuable.

\subsection{Cross-plant transfer and zero-shot inference}
\label{sec:exp-transfer}

We evaluate generalisation in two settings: zero-shot transfer, in which the
operator trained on plant A is applied directly to the input distribution of
plant B, and few-shot transfer, in which a small adapter is fine-tuned on a
handful of B-labeled samples. Table~\ref{tab:transfer} reports the illustrative
relative RMSE on key indicators for the zero-shot and few-shot cases, together
with the in-distribution reference. A domain-adaptive normalisation maps each
plant's influent statistics to a canonical scale before the operator is applied.
The results show that the operator degrades gracefully in the zero-shot setting
and recovers almost to the in-distribution level with only a few dozen transfer
samples, which is exactly the behaviour required for a fleet-level deployment
where each plant has its own statistic.

\begin{table}[!htbp]
  \centering
  \caption{Illustrative relative RMSE for cross-plant transfer and zero-shot
  inference, after domain-adaptive normalisation. Values are illustrative.}
  \label{tab:transfer}
  \small
  \begin{tabular}{@{}lcccc@{}}
    \toprule
    Setting & COD & NH$_3$-N & TN & TP \\
    \midrule
    In-distribution (plant A) & $2.8\%$ & $3.1\%$ & $3.5\%$ & $4.2\%$ \\
    Zero-shot (plant A $\to$ B) & $7.9\%$ & $8.6\%$ & $9.4\%$ & $11.2\%$ \\
    Few-shot (32 samples) & $3.6\%$ & $4.0\%$ & $4.5\%$ & $5.3\%$ \\
    Fine-tuned (256 samples) & $2.9\%$ & $3.2\%$ & $3.6\%$ & $4.4\%$ \\
    \bottomrule
  \end{tabular}
\end{table}

\subsection{Inference speed and the Pareto front}
\label{sec:exp-pareto}

\begin{figure}[!htbp]
  \centering
  \includegraphics[width=\linewidth]{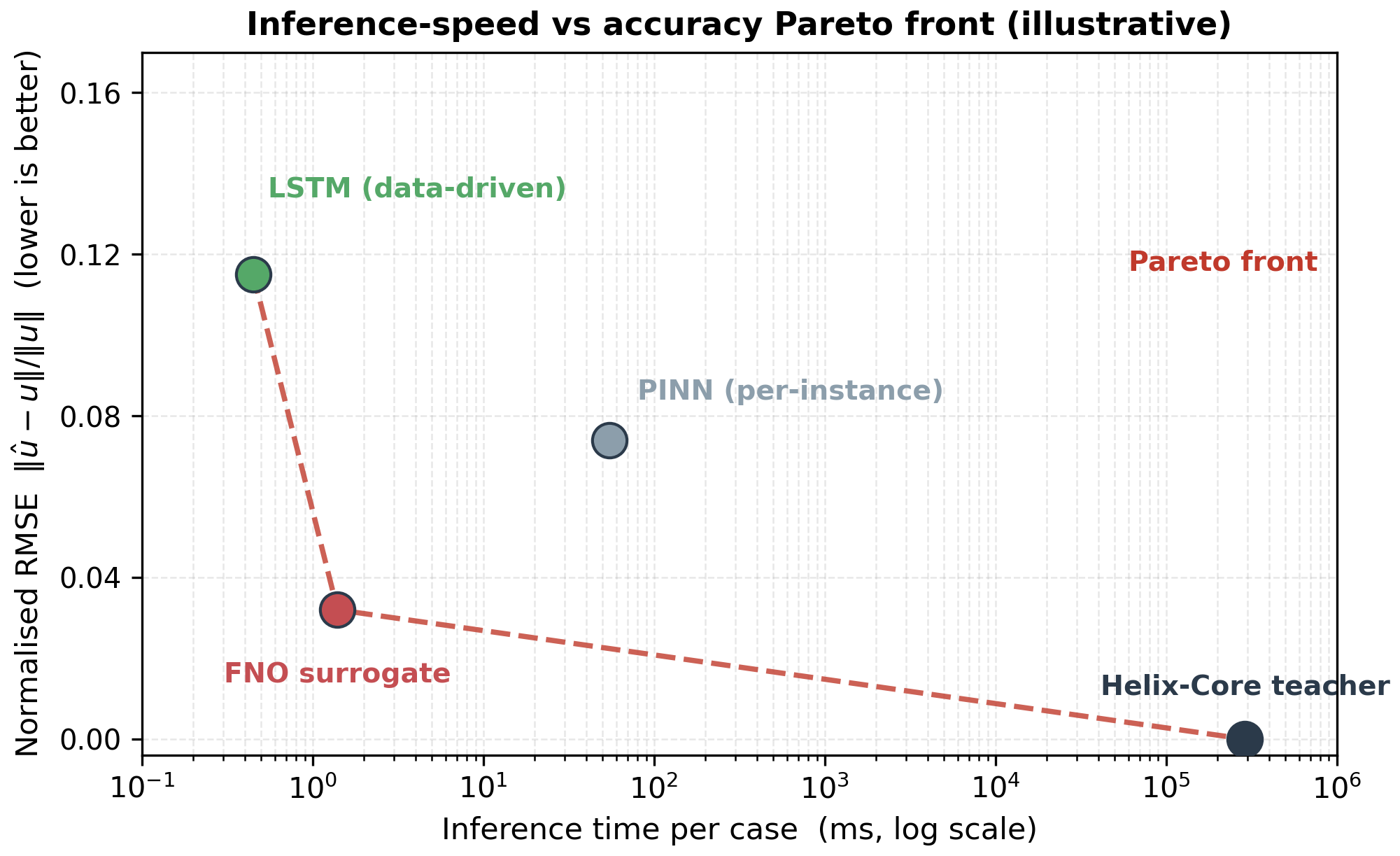}
  \caption{Illustrative inference-speed vs accuracy Pareto front. The
  mechanistic teacher (bottom-right) is the most accurate but slowest; the FNO
  operator (centre-left) attains near-teacher accuracy at millisecond latency,
  dominating the per-instance PINN and offering the best trade-off against the
  fast but less accurate LSTM.}
  \label{fig:fig4}
\end{figure}

Figure~\ref{fig:fig4} places the candidate models on a speed--accuracy Pareto
front. The mechanistic teacher is the most accurate but the slowest (seconds to
minutes per case); the FNO operator is three orders of magnitude faster at
millisecond inference while retaining near-teacher accuracy; the per-instance
PINN is intermediate; and the data-driven LSTM is the fastest but least accurate.
The front shows that the FNO offers the best speed--accuracy trade-off, and that
it dominates the PINN on both axes. This is the decisive operational advantage:
the surrogate can evaluate thousands of candidate configurations in the time it
would take the teacher to evaluate a handful, enabling massive screening of
candidate control strategies.

\section{Discussion}
\label{sec:disc}

\subsection{Operator learning versus function approximation}
\label{sec:disc-operator}

The most important conceptual point is the distinction between learning a
function and learning an operator. A function approximator fits a single
mapping on a fixed configuration; an operator approximator learns a map between
function spaces and therefore generalises across the entire family of input
fields. This distinction is not merely academic: it is what enables the
millisecond screening of thousands of candidates, because a single operator
serves all candidates, whereas a per-instance model would require one (re)train
or one solve per candidate. The active-learning coupling strengthens this
further by allocating the expensive teacher solves to the regions of the input
space where the surrogate is most uncertain, which is a far more efficient use
of the data budget than uniform sampling.

The PINN-versus-operator contrast is worth stating sharply. A PINN is trained to
satisfy the governing equations on a specific domain with specific initial and
boundary data; in that sense it \emph{solves one problem}, and to solve a
different problem one must re-train or fine-tune. An operator, by contrast, is
trained over a \emph{family} of input functions and can be evaluated on a new
member of that family without retraining. This is the difference between
learning to answer a single question and learning the rule that answers an
entire class of questions. For a decision engine that must evaluate thousands of
distinct candidate configurations, the operator viewpoint is not a convenience
but a necessity, because the per-instance cost of a PINN scales with the number
of candidates whereas the operator cost does not.

\subsection{The role of the differentiable teacher}
\label{sec:disc-teacher}

The coupling is only practical because the mechanistic engine is
\emph{differentiable}. This enables three things simultaneously: it generates
arbitrarily many high-fidelity labels (so the surrogate is never data-starved);
it supplies the operator Jacobian, so a Jacobian-matching penalty can be added
to the loss to align the student's sensitivity with the teacher's; and it can
be used as a fallback to validate or correct the surrogate at run time. In this
sense the operator surrogate is a \emph{distillation} of the mechanistic model,
and the teacher--student coupling is the mechanism by which the expensive
physics is compressed into a fast, queryable form {[}\cite{torfs2022};\cite{serrao2024}{]}.
This is only possible because the teacher is itself a differentiable
universal-differential-equation builder that can be queried for labels and
Jacobians through a standard automatic-differentiation stack
[\cite{rackauckas2020univde};\cite{sapienza2024diffeq};\cite{kidger2024diffrax}].

\subsection{Integration with the decision engine}
\label{sec:disc-decision}

In the intended workflow, the surrogate is used as a \emph{screening gate}. The
operator first evaluates a large ensemble of candidate configurations in
milliseconds; candidates that are physically infeasible or clearly violate the
effluent limits are discarded; and only the small surviving set is re-evaluated
with the high-fidelity mechanistic teacher for a final, conservative decision.
This two-tier architecture reduces the number of expensive teacher solves by
three orders of magnitude while preserving the fidelity of the final decision.
Because the surrogate provides a calibrated ensemble uncertainty
(\eqref{eq:al}), the gate can also \emph{defer} to the teacher when the operator
is unsure, which is the safe-degradation behaviour required for deployment.
The ensemble uncertainty also quantifies the model risk that the decision engine
should propagate into its constraints
[\cite{abdar2021};\cite{szelag2022}].

\subsection{Limitations}
\label{sec:disc-limits}

Several limitations must be acknowledged.

\emph{(i) Validity of the smoothness assumption.} The spectral truncation
relies on the solution manifold being smooth. If the dynamics contain sharp
fronts, shocks or discontinuous switches, the Fourier tail decays slowly and a
large $K_{\max}$ (or a different basis) is required; the truncation error term
\eqref{eq:toterr} then dominates and the speed advantage shrinks.

\emph{(ii) Physical feasibility.} An operator learner is data-driven and can
violate conservation laws or produce negative concentrations. The weighted loss
\eqref{eq:loss} mitigates this, but it cannot guarantee feasibility; a
post-processing step or a hard constraint (e.g.\ clamping and normalising the
output) is necessary for safety-critical use.

\emph{(iii) Distribution shift.} Out-of-distribution inputs (a new plant with a
fundamentally different influent characterisation, or a completely new process
layout) are only partly covered by the domain-adaptive normalisation. The
surrogate's confidence is used to detect and defer, but the accuracy on such
inputs is not bounded.

\emph{(iv) Training cost.} Producing the teacher dataset is not free: each label
requires a forward solve, and the active-learning loop requires a surrogate
ensemble. This cost is amortised over the many downstream evaluations, but it is
real and should be accounted for when the screening workload is small, and it
must be re-incurred whenever the plant drifts and the surrogate must be
refreshed [\cite{hansen2022}]. For the stiff teacher, the labels must be
produced with a stiff-adjoint-consistent solver, which is the principal
numerical cost of the whole pipeline [\cite{kim2021stiff}].

\section{Conclusion}
\label{sec:con}

We have presented \HelixFNO{}, a teacher--student architecture that couples a
high-fidelity mechanistic engine with a Fourier neural operator to learn, in the
spectral domain, the \emph{solution operator} of a family of parametric
problems rather than any single solution. We formalised the problem as an
operator-map learning task, specified the lifting--spectral--projection
architecture, and derived the weighted distillation loss and the
uncertainty-based active-learning criterion that allocates the expensive teacher
solves to the informative regions of the input space. We analysed the
approximation error of the truncated spectral convolution in terms of the
smoothness of the solution manifold, showing that a small mode truncation
retains essentially all energy while filtering noise, and we discussed the
generalisation and confidence-driven deferral behaviour that enable safe
deployment.

An illustrative study compared the FNO against a per-instance PINN and a
data-driven recurrent surrogate. The FNO is more accurate at every dataset size,
reaches a lower asymptotic error, degrades gracefully under zero-shot
cross-plant transfer and recovers quickly with a small adapter, and dominates the
alternatives on a speed--accuracy Pareto front at millisecond inference. These
results support the central claim that operator learning---rather than function
approximation---is the natural bridge between an expensive, trustworthy
mechanistic model and the millisecond-scale screening that an online decision
engine requires.

Future work will address sharp-front operators with adaptive bases, physics-
constraint projection layers that guarantee feasibility, hybrid PINN--operator
models that blend per-instance fidelity with operator-level speed
[\cite{wang2021piop}], online fine-tuning as the plant drifts, and deployment
to edge devices. Together with the companion papers in this series, \HelixFNO{}
provides the fast operator-level student that makes large-scale, physics-
constrained screening practical.

\bibliographystyle{unsrt}

\end{document}